\documentclass[a4paper]{article}

\usepackage[top=2.5cm, bottom=2.5cm, left=2.5cm, right=2.5cm]{geometry}

\usepackage[pdftex,colorlinks=true,linkcolor=red,citecolor=blue,urlcolor=green,bookmarks=false,pdfpagemode=None]{hyperref}

\usepackage{mathptmx}
\usepackage{graphicx}
\usepackage{arabtex}
\usepackage{CJKutf8}
\usepackage{verbatim}
\usepackage{alltt} 
\usepackage{multicol}

\usepackage{natbib} 

\newcommand{\JP}[1]{%
\begin{CJK}{UTF8}{min}%
#1
\end{CJK}%
}

\begin{document}

\title{Sentence Object Notation: \\Multilingual sentence notation based on Wordnet}

\author{Abdelkrime Aries \and Djamel eddine Zegour \and Walid Khaled Hidouci}
\date{\textit{Ecole nationale Sup\'erieure d'Informatique (ESI, ex. INI), Algiers, Algeria} \\[.2cm]
	Emails: \{ab\_aries, d\_zegour, w\_hidouci\}@esi.dz}

\maketitle

\begin{abstract}
The representation of sentences is a very important task. 
It can be used as a way to exchange data inter-applications. 
One main characteristic, that a notation must have, is a minimal size and a representative form.
This can reduce the transfer time, and hopefully the processing time as well.

Usually, sentence representation is associated to the processed language. 
The grammar of this language affects how we represent the sentence.
To avoid language-dependent notations, we have to come up with a new representation which don't use words, but their meanings. 
This can be done using a lexicon like wordnet, instead of words we use their synsets.
As for syntactic relations, they have to be universal as much as possible.

Our new notation is called STON "\textit{SenTences Object Notation}", which somehow has similarities to JSON. 
It is meant to be minimal, representative and language-independent syntactic representation.
Also, we want it to be readable and easy to be created. 
This simplifies developing simple automatic generators and creating test banks manually.
Its benefit is to be used as a medium between different parts of applications like: text summarization, language translation, etc. 
The notation is based on 4 languages: Arabic, English, Franch and Japanese; and there are some cases where these languages don't agree on one representation. 
Also, given the diversity of grammatical structure of different world languages, this annotation may fail for some languages which allows more future improvements. 
\end{abstract}


\section{Introduction}
\label{sec:intro}

Tagging sentences is a very important task in natural language processing.
One of the famously known methods is syntactic tagging. 
The main idea is to detect the different parts (structure) of a sentence, such as nominal phrases, verbal phrases, nouns, verbs, etc.
This structure can be expressed using some languages like XML, JSON, etc. \citep{2008-bertran-al,2009-recasens-marti,2011-lopatkova-al}.
The problem with syntactic tagging is its dependency to the processed language. 
Indeed, it is a good way if our system is destined for a specific language. 
But, when it comes to multilingual or cross-lingual systems, it is better to come up with a way to represent the sentence structure independently from languages. 

Semantic representation of sentences is one solution to this problem. 
The meaning of a sentence is something beyond languages; it is related to the different concepts of its words and the different relations between them. 
Words are just a way to describe a concept in a given language.
For example, the words "\RL{^sajaraT} \textit{/shajarah/}" in Arabic, "\textit{tree}" in English, "\textit{arbre}" in French and "\JP{木}\textit{ /ki/}" in Japanese refer to the same concept. 
This concept in Wordnet 3.0\footnote{WordNet is a large lexical database of English. Nouns, verbs, adjectives and adverbs are grouped into sets of cognitive synonyms (synsets), each expressing a distinct concept. URL: \url{https://wordnet.princeton.edu}}\citep{1995-miller:1995} is defined as "\textit{a tall perennial woody plant having a main trunk and branches forming a distinct elevated crown; includes both gymnosperms and angiosperms}". 
The idea of semantic representation is to represent the words as concepts and each language can link its own words to these concepts. 
Then, the semantic relations between these concepts are extracted from the sentence \citep{1999-uchida-al,2013-banarescu-al}.
It will be great if we can gather all concepts of different languages, then make links to each one of them. 
Also, to detect the semantic relations between the concepts in a sentence automatically, that will need a large amount of knowledge and processing power. 

Our idea is to propose a simple language which will help us transfer the information in sentences between applications. 
For example, we can use it to send sentences from a summarization system to a translation system. 
That allows us to create what is known as cross-lingual summarization. 
The language has to be simple and unambiguous so the developers can, easily, create tools to encode a specific natural language to this one. 
It must have a minimal size to minimize the transfer time and also the processing one. 
One of the main characteristics on which we insist is the readability, which helps us to create STON representations manually. 
To this end, we want to represent the syntactic relations in sentences taking in mind the multilingualism aspect.
We must insist on the term \textbf{multilingual} which means "\textit{using several languages}", so it doesn't have to represent \textit{all} world languages.
In our study, we are interested in four languages: Arabic, English, French and Japanese. 
For non Latin scripts, namely Arabic and Japanese, we will provide the ALA-LC romanization\footnote{ALA-LC (American Library Association - Library of Congress) is a set of standards for romanization to represent texts in other writing systems using the Latin script. URL: \url{https://www.loc.gov/catdir/cpso/roman.html}}.

The rest of this paper is organized as follows. 
Section \ref{sec:related} presents some related words in the subject of text annotation. 
Section \ref{sec:structure} describes our main proposition and different parts of STON. 
Section \ref{sec:compl-str} addresses the cases where we have adpositional phrases, relative clauses and comparison. 
In section \ref{sec:issues}, some solutions are proposed concerning the coordination between references, proper names, verbs with two objects, complementizer and passive voice. 
Section \ref{sec:discussion} is reserved to discuss our work, its grounds, its contribution in comparison with other works, its benefits and its limits.
Finally, section \ref{sec:conclusion} is reserved for conclusion and future improvements.

\section{Related works}
\label{sec:related}

Sentence structure can be represented using generic-purpose languages such as XML or JSON. 
Lets take XML as an example, to represent a sentence we have to specify some roles in a DTD file. 
It would contain different structures of a sentence, such as subject, object, verb, tense of the verb, etc.
For example, \citet{2009-recasens-marti} uses XML to represent sentences in order to annotate corpora for Spanish and Catalan. 
They used AnCoraPipe \citep{2008-bertran-al} to create the corpus.
Figure \ref{fig:recasens-marti} represents an example of XML annotation of the sentence "\textit{La Comisi\'{o}n Europea anunci\'{o} \ldots}" which means "\textit{The European Commission announced \ldots}".
They use XML tags to express nouns, verbs, nominal phrases, etc. where each tag has some properties. 

\begin{figure}[ht]
\begin{tabular}{|c|}
\hline
\\
\input{Codes/recasens-marti.tex}
\\
\\
\hline
\end{tabular}
\caption{Example of XML sentence annotation \citep{2009-recasens-marti}}
\label{fig:recasens-marti}
\end{figure}

%

Knowledge-based interlingual machine translation uses a representation of sentences as medium between the source language and the destination one.
KANT \citep{1991-mitamura-al} is an example system that uses an interlingua to represent sentences before translation.
KANT interlingua is a list-based structural representation scheme using nested frames.
Each interlingua frame contains a head concept, series of feature-value pairs, and semantic slots containing nested interlingua frames.
Concepts are symbols which begins with an asterisk (*) followed by a concept prefix defining its category (e.g. *A-DRIVE which is the action drive).
KANT interlingua distinguishes many categories such as action, object, manner, proper name, etc.
It uses certain features of the input text, such as modality, aspect, discourse markers, etc. in order to generate grammatically accurate output texts \citep{1991-mitamura-al}.
Semantic roles are relations between frames, such as agent, theme, etc.
Figure \ref{fig:kant} represents the KANT interlingua's representation of the sentence "\textit{The default rate remained close to zero during this time.}".
It contains concepts such as *A-REMAIN, *K-DURING, etc.; features such as FORM, TENSE, etc. and semantic roles such as Q-MODIFIER, THEME, etc.

\begin{figure}
	\centering
\begin{tabular}{|c|}
\hline
\\\small
\input{Codes/kant.tex}
\\
\\
\hline
\end{tabular}
\caption{Representing the sentence "\textit{The default rate remained close to zero during this time.}" in KANT interlingua \citep{1998-mitamura-al}}
\label{fig:kant}
\end{figure}

Despite being an interlingua, KANT shows some limitations when representing sentences \citep{1998-mitamura-al}.
It is more close to English semantics than being multilingual, because it is intended for English to other languages translation.
Also, It is designed for technical domains, thus the vocabulary is limited to a subset of meanings.

Universal Networking Language (UNL) is a knowledge representation language to represent the meaning of texts without ambiguity. 
It was developed in 1996, as un intermediate multilingual language to be used through the Internet \citep{1999-uchida-al, 2005-uchida-zhu}.
The major commitments of the UNL are the following:
\begin{itemize}
\item It must represent information: represent "what was meant" and not "what was said" or "how it was said".
\item It must be a language for computers: like HTML, SGML, XML, etc.
\item It must be self-sufficient: The UNL representation must not depend on any implicit knowledge and should explicitly codify all information.
\item It must be general-purpose: Its primary objective is to serve as an infrastructure for handling knowledge. 
It can be used for different tasks such as: translation, text mining, multilingual document generation, summarization, etc.
\item It must be independent from any particular natural language.
\end{itemize}
UNL defines some tags for the structure of the text: document "\textbf{[D]}", paragraph "\textbf{[P]}" and sentence "\textbf{[S]}".
Concepts are represented as character-strings called "\textbf{Universal Words (UWs)}", where each natural language (En, Fr, etc.) has its own word dictionary.
UNL expressions are based on binary relations, where each binary relation has two UWs as parameters.
Also, UNL specifies some attributes to represent information conveyed by natural language grammatical categories (such as tense, mood, aspect, number, etc).
Figure \ref{fig:unl} represents UNL formulation of the sentence "\textit{Human affect the environment}".
The sentence starts with the tag "\textbf{\{unl\}}" followed by two binary relations. 
The relation "\textbf{agt}" defines a thing which initiates an action. 
In our example, "Human" which is a plural noun is the one who do (in present tense) the action of "\textit{affecting}".
The relation "\textbf{obj}" defines a thing in focus which is directly affected by an event or state.
In our example, "\textit{environment}" is the direct object of the action of "\textit{affecting}".

\begin{figure}
	\centering
\begin{tabular}{|c|}
\hline
\\
\input{Codes/unl.tex}
\\
\\
\hline
\end{tabular}
\caption{Representing sentence "\textit{Human affect the environment}" in UNL}
\label{fig:unl}
\end{figure}

Indeed, UNL has a minimal size, with a multilingual background and covers a large number of languages.
However, it shows some limitations when it comes to which relations we must choose. 
\citet{2005-boguslavsky} claims that the selection of relations differs from team to team as it is sometimes ambiguous which one to choose.
An example of that is the phrase "freedom for all" which was described with the purpose relation "\textbf{pur}" and with the beneficiary relation "\textbf{ben}" by another team.
\citet{2013-martins} raises some other issues concerning UNL. 
One of these issues is the proper nouns, are they treated as permanent UWs or just temporary? 
Also, a concept can be represented by simple UW or a compound UW.
For example "\textit{the physiological need for food}" can be represented using the UW "\textbf{hunger}", as it can be represented with the compound UW "\textbf{hungry.@ness}".
Technically speaking, we can say it is hard to design an encoder from natural language to UNL. 
After parsing a sentence, we have to find the relations between each part of this sentence. 
It is, sometimes, too ambiguous to select between two relations manually, let alone selecting them automatically.

\begin{figure}
	\centering
\begin{tabular}{|ccc|}
\hline
&&\\
\input{Codes/amr-log.tex}
 &
\input{Codes/amr.tex}
 &
\begin{minipage}{.42\textwidth}
     \includegraphics[width=4cm]{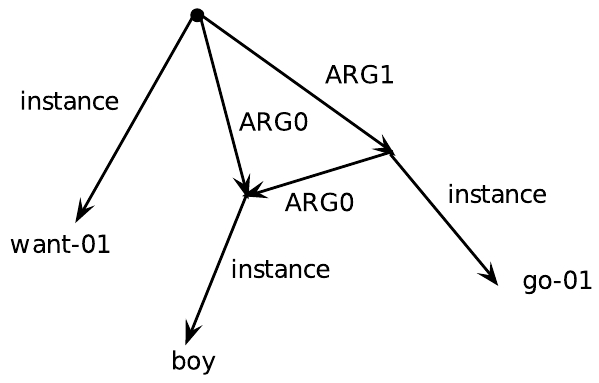}
\end{minipage}\\
&&\\
LOGIC format & AMR format & GRAPH format \\
\hline
\end{tabular}
	\caption{AMR representation of the sentence "\textit{The boy wants to go}" \citep{2013-banarescu-al}.}
	\label{fig:amr}
\end{figure}

A most recent representation language is AMR (Abstract Meaning Representation) proposed by \citet{2013-banarescu-al}.
It is a semantic representation language designed to represent the meaning of English sentences. 
The text is represented as a graph, where the leaves are labeled with concepts such as "\textbf{(b / boy)}" which means an instance called "b" of the concept "boy" (See Figure \ref{fig:amr}).
The concepts are English words ("boy"), PropBank framesets \citep{2002-kingsbury-palmer} ("\textbf{want-01}”), or special keywords: special entity types ("\textbf{date-entity}", "\textbf{world-region}", etc.), quantities ("\textbf{monetary-quantity}", "\textbf{distance-quantity}",etc.) and logical conjunctions ("\textbf{and}", etc).
The relations between concepts are:
\begin{itemize}
\item  Frame arguments, following PropBank conventions.
\item General semantic relations.
\item  Relations for quantities.
\item Relations for date-entities.
\item Relations for lists.
\end{itemize}

\noindent

AMR is light weight annotation, but it is heavily based on English.
It has some limitations where it comes to inflectional morphology for tense and number.
It does not deeply capture many noun-noun or noun-adjective relations.
Also, because it relies on Propbank framesets, it is subject to the  Propbank constraints.

\section{STON representation}
\label{sec:structure}

\begin{figure}
	\centering
\includegraphics[width=.8\textwidth]{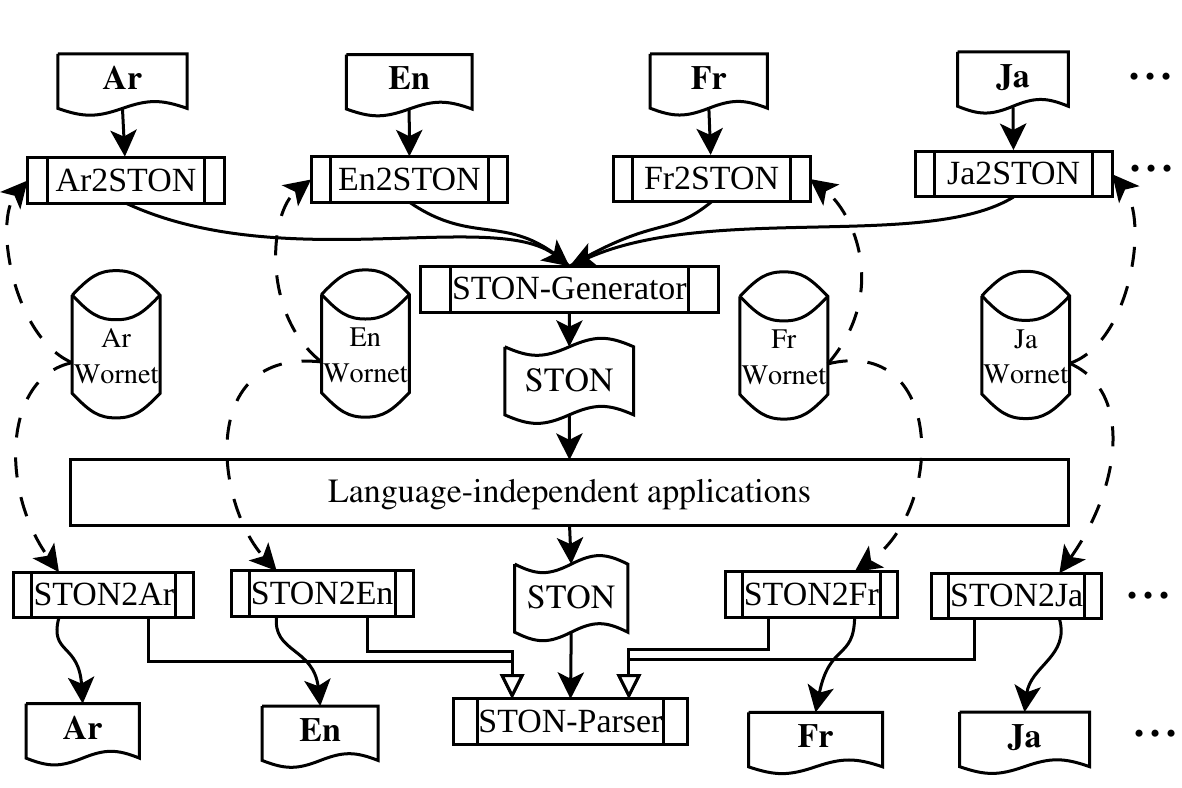} 
\caption{Example of STON's purpose}
\label{fig:lang-ston}
\end{figure}

STON is intended to represent sentences' syntactic structures in a multilingual context.
Figure \ref{fig:lang-ston} represents our vision on how STON may be used.
We can use lexical parsers of several languages to get the different parts of speech. 
Then, we can use them to create STON representation using a STON generator and Wordnet lexicon of each language. 
The representation can be used in multiple multilingual applications, like the example we mentioned earlier. 
Then having a STON parser, we can extend it to handle each language apart. 
Using a realizer or a language generator, we can reproduce a readable text for the destination language.

In our representation, we look at different parts of sentences as either actions or roles. 
Actions are the dynamic part of the sentence.
Each action contains the verb and its morphological specifications like tense, negation, etc.
While roles are, generally, nominal phrases that have a purpose in the action.
They can be agents, themes, places, times, etc. 

\subsection{Roles}
\label{subsec:roles}

Nominal phrases play roles in an action; They can be agents, themes, places, times, etc. 
In our case, agents are those who do the action (either in active or passive voice), and themes are those who undergo or experience the action.
For example, in table \ref{tab:role-action}, "\textit{the fat man}" and "\textit{delicious food}" are roles played in the action of "\textit{eating}" which happened in the present.
The first phrase plays the role of an agent, while the second plays the role of a theme.
\begin{table}
	\centering
\caption{Roles and Actions in Arabic, English, French and Japanese. \label{tab:role-action}}%
\begin{tabular}{ll}
\hline\noalign{\smallskip}
Language & Sentence \\
\noalign{\smallskip}\hline\noalign{\smallskip}
Arabic & \RL{alrrajulu alssamiynu ya'ku--lu al-la.hma al-la_diy_d.} \\
& /al-Rajulu al-Sam\={\i}nu ya'kulu al-La\d{h}ma al-Ladhidh./ \\
English & The fat man eats the delicious meat. \\
French & Le gros homme mange la viande d\'eliciouse. \\
Japanese & \JP{太った男性は美味しい肉を食べます。} \\
& /futotta dansei wa oishii niku o tabemasu./ \\
\noalign{\smallskip}\hline
\end{tabular}

\end{table}

In our representation (STON), we list all the roles that appear in the sentences. 
Each role contains the following attributes (see Figure \ref{fig:roles}):
\begin{itemize}
\item \textbf{id:} a name for the role in order to reference it.
\item \textbf{syn:} the synset number of the noun in the lexicon (Wordnet in our case). 
\item A set of adjectives blocks that modify the noun.
Each adjective block contains its synset and a set of adverbs synsets.
\item \textbf{qnt:} the quantity; it describes the amount of the noun. 
For example "\textit{10 apples}", the quantity is 10.
By default, it equals 1; it is a number but it can have the term "\textbf{PL}" for plural.
Some languages, like Arabic, has dual numbers "\RL{mu_tannY} \textit{/muthanna/}" which can simply represented by the number 2.
Then, the language generator can handle this.
As for ordinal numbers, we add "\textbf{O}" before the number; example, "\textbf{O2}" means second.
\item \textbf{def:} defined; many languages, such as Arabic, English and French, have the ability to identify a noun from many using the definite articles.
\end{itemize}

\begin{figure}
	\centering
\begin{tabular}{|l|}
\hline
\\
\input{Codes/roles.tex}
\\
\\
\hline
\end{tabular}
	\caption{Roles representation.}
	\label{fig:roles}
\end{figure}

\subsection{Actions}

The verb in a sentence, or in a clause, represents an action. 
In languages, like Arabic, there are some nominal sentences, which doesn't include a verb. 
Nevertheless, we can add a verb to express the action, like the examples in Table \ref{tab:ar-nom-sent}. 
The first sentence is composed of a subject "\RL{mubtada'} \textit{/mubtada'/}" (which is a Noun) and a predicate "\RL{xabar} \textit{/khabar/}" (which is an Adjective). 
In the second one, the subject is a noun and the predicate is a prepositional phrase. 
In both sentences, we can use the copula (to be) as if it is the action, the subject will be considered as an agent and the predicate as a theme.
In case of the third sentence, where we have an active participle "\RL{'ism faA`il} \textit{/ism f\=a`il/}", and the origin verb is a movement verb like "\textit{to go}" we can consider it as present continuous \citep{1997-haak}.
\begin{table}
	\centering
\caption{Example of nominal sentences in Arabic. \label{tab:ar-nom-sent}}%
\begin{tabular}{lll}

\hline\noalign{\smallskip}
 & Arabic & English \\
 
\noalign{\smallskip}\hline\noalign{\smallskip}
Noun + & \RL{alrrajulu (yakuwnu) samiynuN.} & The man (is) fat. \\
Adjective & /al-Rajulu (yak\=unu) sam\={\i}nun./ & \\

\noalign{\smallskip}\hline\noalign{\smallskip}
Noun + Prep. & \RL{alrrajulu fiy al-ssuwqi.} & The man (is) in the market. \\
+ Noun & /al-Rajulu f\={\i} al-S\={u}qi./ & \\

\noalign{\smallskip}\hline\noalign{\smallskip}
Noun + Active & \RL{alrrajulu \textbf{_daAhibuN} 'ilY al-ssuwqi.} & The man is going to the market. \\
participle + PP & /al-Rajulu dhahibun il\'{a} al-S\={u}qi./ & \\

\noalign{\smallskip}\hline

\end{tabular}

\end{table}

To represent the actions, we define a set of actions blocks, where each block contains some attributes. 
The attributes are generally related to the verb, since the action is all about the verb. 
But, it contains also some links to agents, themes, etc. 
These are the main attributes contained in each action, where "id" and "syn" are compulsory attributes (See Figure \ref{fig:actions}):
\begin{itemize}
\item \textbf{id:} a name for the action in order to reference it.
\item \textbf{syn:} the synset number of the verb in the lexicon (Wordnet). 
\item \textbf{agt:} a set of role IDs; those who did the action.
\item \textbf{thm:} a set of role IDs; those who receive the action.
\item A set of adverbs blocks that modify the verb.
Each adverb block contains its synset and a set of adverbs synsets.
\item Some verb specifications: tense ("\textbf{tns}"), progression ("\textbf{prg}"), perfect aspect (\textbf{prf}), negation ("\textbf{neg}") and modality ("\textbf{mod}").
\end{itemize}

\begin{figure}
	\centering
\begin{tabular}{|c|}
\hline
\\
\input{Codes/actions.tex}
\\
\\
\hline
\end{tabular}
	\caption{Actions representation.}
	\label{fig:actions}
\end{figure}

The tense can be: past ("\textbf{PA}"), present ("\textbf{PR}") or future ("\textbf{FU}").
The absence of tense means the action is tense-free; e.g. "\textit{To do so, you must try}".
There are languages which doesn't have future tense, such as Arabic and Japanese.
In this case, we can detect the tense using adverbs ("\textit{Tomorrow}") or temporal prepositional phrases ("\textit{in the next year}").
Other languages define (far/near) past and (far/near) future. 
For example, in Arabic, there is no tense called "future" but it can be expressed using auxiliaries. 
For near future, we use "\RL{sa-} \textit{/sa/}" attached to the verb in present tense ("\RL{sa'a_dhabu} \textit{/sa'dhhabu/}", I will go \textless soon\textgreater). 
For far future, we use "\RL{sawfa} \textit{/sawfa/}" detached from the verb in present tense ("\RL{sawfa 'a_dhabu} \textit{/sawfa adhhabu/}", I will go \textless later\textgreater).
But, since this can be detected using adverbs such as "\textit{soon}" and "\textit{later}", we can ignore it.

There are two aspects which are mostly used in occidental languages: progressive and perfect. 
The perfect aspect refers to some actions prior to the time under consideration, which is viewed as already completed. 
In STON representation, the tense is imperfect unless we add ("\textbf{prf: Y;}").
As for progressive aspect, it is a situation where a verb is (was) in motion for an interval of time. 
Likely, the action is not progressive unless we add (\textbf{prg: Y;}).

Modality, in our case, can express possibility ("\textbf{MAY}"), admissibility ("\textbf{CAN}") or obligation ("\textbf{MUST}"). 
The modal verb "\textit{will}" is used to express the future, which is a tense in our case. 
Advice ("\textit{You should see a doctor}"), prohibition ("\textit{You mustn't smoke here}"), certainty ("\textit{He must be rich, since he lives there}"), permission ("\textit{You can leave now}") and lack of necessity ("\textit{You don't have to do anything}") can be represented using these three modal verbs.

\subsection{Sentences}

A Role-Action representation is not sufficient, since there are sentences which contains consecutive actions.
In Table \ref{tab:cons-act}, we can observe three actions: going to the market, coming back and watching T.V.
If we represent this sentence as three sentences instead, we will loose the information that these actions are consecutive.
Not to mention, we have to specify which actions are the main ones (there are actions which are relatives of roles and actions).

\begin{table}
	\centering
\caption{Example of Consecutive actions. \label{tab:cons-act}}%
\begin{tabular}{ll}
\hline\noalign{\smallskip}
Language & Sentence \\
\noalign{\smallskip}\hline\noalign{\smallskip}
Arabic & \RL{_dahaba kariym 'ilY as-suwqi _tumma `aAda 'ilY al-manzili _tumma ^saAhada at-tilfaAza.} \\
& /dhahaba Kar\={\i}m il\'{a} al-S\={u}qi thumma `\={a}da il\'{a} al-manzili thumma sh\={a}hada al-Tilf\={a}za./ \\
English & Karim went to the market, came back home, then watched T.V. \\
French & Karim est allé au marché, il est revenu à la maison, ensuite il a regardé la télé. \\
Japanese & \JP{カリムさんは市場に行って、家に戻って、テレビを目ました。} \\
& /Karimu-san wa itchiba ni itte, ie ni modotte, terebi o memashita./ \\
\noalign{\smallskip}\hline
\end{tabular} 

\end{table}

The sentence part lists some sentences blocks.
Each block contains the type of the sentence: affirmation ("\textbf{AFF}"), exclamation ("\textbf{EXC}"), question ("\textbf{QST}") and imperative ("\textbf{IMP}"). 
It have an attribute "\textbf{act}" to list the references of actions in this sentence.
The annotation of sentences in STON is illustrated in Figure \ref{fig:sent}.

\begin{figure}
	\centering
\begin{tabular}{|c|}
\hline
\\
\input{Codes/sent.tex}
\\
\\
\hline
\end{tabular}
	\caption{Sentences representation in STON.}
	\label{fig:sent}
\end{figure}

\section{Relations}
\label{sec:compl-str}

\begin{figure}
	\centering
\includegraphics{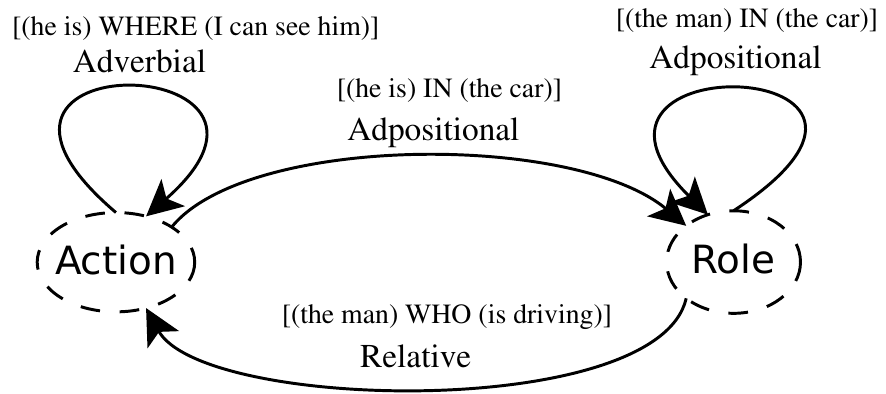} 
\caption{Relations between roles and actions}
\label{fig:relations}
\end{figure}

There are many relations between the clauses and phrases of a sentence.
Attributes such as agents and themes in an action can just represent simple sentences. 
A more complicated sentence can contain adpositional phrases, relative clauses, etc.
Figure \ref{fig:relations} shows our view about the 4 relations between the roles and the actions.

A Role-Role relation can be expressed by adpositional relations as the example "\textit{the man in the car}". 
Adpositional relations can, also, be used to express Action-Role relation ("\textit{he is in the car}").
Another relation is the relation Role-Action found in relative clauses ("\textit{the man who is driving}").
Like adjectives, a relative clause can modify (describe) a nominal phrase. 
A noun can be described by a relative clause, such in the sentence "\textit{the man who was strong}".
The late example is different from "\textit{the strong man}" because the first sentence adds the information of a past quality of strength. 
The relation Action-Action can be found in adverbial clauses (\textit{he is where I can see him}).
Comparison between two roles is another issue; Are they equal or one is more or less than the other?
They can share a verb such in "\textit{I do more work than you}" or an adjective "\textit{I am stronger than you}".
More structure have to be added in order to allow these types of relations.

\subsection{Relative clauses}

Relative clauses are a little challenging to be represented using our Role-Action notation. 
Our notation starts with roles then actions. 
Since relative clauses are actions which describe a role, they have to be represented in the role itself.
In Table \ref{tab:rel-cl-eg}, the man drinking water, is the one who ate the meat earlier. 

\begin{table}
	\centering
\caption{Example of relative clauses. \label{tab:rel-cl-eg}}%
\begin{tabular}{ll}
\hline\noalign{\smallskip}
Language & Sentence \\
\noalign{\smallskip}\hline\noalign{\smallskip}
Arabic & \RL{alrrajulu alla_diy 'akala al-la.hma ya^srabu almaA'a.} \\
& /al-Rajulu al-Ladh\={\i} akala al-La\d{h}ma yashrabu al-M\={a}'./  \\
English & The man who ate the meat is drinking water. \\
French & L'homme qui a mang\'e de la viande boit de l'eau. \\
Japanese & \JP{肉を食べた男性が水を飲んでいます。} \\
& /niku o tabeta dansei ga mizu o nonde imasu./ \\
\hline\noalign{\smallskip}
\end{tabular}

\end{table}

One proposition is to use ulterior referencing in roles to reference an action as a relative. 
Then, the relationship between the role and the relative clause must be specified. 
For example, the text in Table \ref{tab:rel-cl-eg} can be represented as in Figure \ref{fig:rel-cl-eg}.

\begin{figure}
	\centering
\begin{tabular}{|c|}
\hline
\\
\input{Codes/rel-cl-eg.tex}
\\
\\
\hline
\end{tabular}
	\caption{Example of STON representation in case of relative clauses.}
	\label{fig:rel-cl-eg}
\end{figure}

Looking at relative pronouns (See Table \ref{tab:rel-pron}), we can consider 4 main types of relative phrases: subject, possessive, direct object and indirect object. 
In this categorization, we considered a person as same as a thing, because the difference between them can be taken in consideration in text generation task.

\begin{table}
	\centering
\caption{Relative pronouns.\label{tab:rel-pron}}%
\begin{tabular}[b]{llllll}
\hline\noalign{\smallskip}
& Person & thing & Place & Time & Reason \\
\noalign{\smallskip}\hline\noalign{\smallskip}
Subject & who/that & which /that & & & \\
Object & who/whom/that & which/that & where & when & why \\
Possessive & whose & whose &&& \\
\noalign{\smallskip}\hline
\end{tabular}

\end{table}

\begin{itemize}
\item The subject type ("\textbf{SBJ}") indicates that the main clause is a subject of the relative one. 
For example, in the phrase "\textit{the man who ate the meat}", the main clause "\textit{the man}" is the subject of the relative one "\textit{ate the meat}".

\item The possessive type ("\textbf{POS}") indicates that the noun in the relative clause is possessed by the main one.
For example, the phrase "\textit{the man whose car is so expensive}" describes a man who possess an expensive car.

\item The object type ("\textbf{OBJ}") indicates that the main clause is a direct object of the relative one.
For example, the phrase "\textit{the man whom I saw yesterday}" describes a man who is an object of me seeing him.
We can deduce the following information from it "\textit{I saw \textbf{the man} yesterday}".

\item The indirect object, or the clauses that starts with prepositions, are various. 
The meaning of these clauses follows the meaning of their prepositions. 
In this case, we can form their types using the keyword ("\textbf{IO\_}") followed by the Adpositional relation (next subsection).
For example, "\textit{The town from which I came}" describes a town which is an indirect object of me coming.
We can deduce the following information from it "\textit{I came \textbf{from the town}}".
As for the relative adverbs "\textit{where}" and "\textit{when}", we can replace them with a preposition. 
For example, "\textit{The town where I met him}" is equivalent to "\textit{The town in which I met him}".

\item One other type is the reason ("\textbf{RSN}"), expressed by the relative adverb "\textit{why}".

\end{itemize}

\subsection{Adpositional phrases}

An adpositional phrase includes prepositional phrases, postpositional phrases, and circumpositional phrases. 
Usually, they are used to describe the time or the place of an action. 
Arabic, English and French use prepositional phrases, while Japanese uses postpositional phrases. 
They can be related to an action: "He works at 8", or a role: "The mother of the boy".

Our objective is to define as few relations as possible which can represent the meaning of most adpositions.
For that matter, we define these relations, which can be extended in the future:
\begin{itemize}
\item \textbf{AGO}: An amount of time back in the past. 
e.g. "\textit{I bought it 2 years ago}".
\item \textbf{FRM}: An origin or change of state; 
e.g. "\textit{I came from Algeria}".
\item \textbf{IN}: The existence of a particular time, place or situation.
e.g. "\textit{I wake up at 7 am.}", "\textit{I was born in 1986}", "\textit{I am in town.}", "\textit{I am at home.}"
\item \textbf{SNC}: A period from past till a particular time. 
e.g. "\textit{I have worked since 2013}".
\item \textbf{TO}: A destination; It can be a location, a person, etc. 
e.g. "\textit{I am going to the market.}", "\textit{I will give it to him.}","\textit{We waited till noon}".
\item \textbf{FOR}: An amount of time or an objective. 
e.g. "\textit{I will sleep for 2 hours}", "\textit{the foundation for the support of the cinema.}", "\textit{during the war}".
\item \textbf{BEF}: A time earlier than another (before) or a place in front of something.
 e.g. "\textit{She's always up before down}", "\textit{I sit in front the TV.}".
\item \textbf{AFT}: A time further than another (after) or a place at the back of something (behind). 
e.g. "\textit{He always sleep after 10 pm}", "\textit{the towel is behind the door}".
\item \textbf{BY}:  Not later than a specified time, or a place besides something or an agent; 
 Here, we consider these prepositions as the same: "\textit{by}", "\textit{next to}", "\textit{besides}" and "\textit{near}".
e.g. "\textit{I will finish by 5pm.}", "\textit{He walks by the river.}", "\textit{diffraction by crystals}". 
\item \textbf{INS}: Something inside something else.
e.g. "\textit{The present is inside the box.}".
\item \textbf{OUT}: Something outside something else.
e.g. "\textit{I am outside the home}".
\item \textbf{BLW}: A place under something; Here, we treat "\textit{Below}" and "\textit{Under}" as the same.
e.g. "\textit{He swims under the bridge.}".
\item \textbf{ABV}: A place above something; Here, we treat "\textit{Above}" and "\textit{over}" as the same.
e.g. "\textit{we walked over the bridge.}".
\item \textbf{BTW}: A place between two or more things. 
e.g. "\textit{The museum is between two flats.}".
\item \textbf{THR}: From one side to another, or surrounded;
Here, we don't make a difference between "\textit{through}" and "\textit{across}". 
e.g. "\textit{She walked through the forest.}".
\item \textbf{ON}: A subject or something connected to something else. 
e.g. "\textit{She borrowed a book about mathematics}".
\item \textbf{WTH}: Being together or being involved.
e.g. "\textit{I am with him}".
\item \textbf{OF}: Possession or belonging. 
e.g. "\textit{The leafs of the tree}".
This relation is used to represent compound nouns: "\textit{printer cartridge}" will be "\textit{cartridge of printer}".
\item \textbf{AS}: A role. 
e.g. "\textit{He works as a teacher.}".
\item \textbf{UND}: A situation. 
e.g. "\textit{He did the exam under a lot of pressure.}".
\end{itemize}

In English, "\textit{at}" and "\textit{in}" are used to express "\textit{exact}" and "\textit{wide}" locations or times respectively.
In Japanese, there is no difference between exact or wide range locations and times.
But, there is another aspect which is used to distinguish the prepositions "\JP{に} \textit{/ni/}" and "\JP{で} \textit{/de/}". 
The first one (/ni/) is used to express the existence of something or someone in a place.
e.g. "\JP{犬は公園にいます。} \textit{/inu wa k\={o}en ni imasu./}" ("\textit{The dog is in the park.}").
While the second one (/de/) is used to express the place where an action takes place.
e.g. "\JP{犬が公園で吠えます。}" /inu ga k\={o}en de hoemasu./ (The dog barks in the park.).
Unfortunately, our representation doesn't take these aspects into consideration.
Nevertheless, we believe this can be solved in text generation task.
In case of "at-in" differentiation, we can test the noun to generate the right preposition.
Likewise, for Japanese postpositions "ni-de", we can use the verb.

\subsection{Adverbial clauses}

An adverbial clause is a dependent clause which functions as an adverb. 
in our representation, it represents the relation between an action and another.
These are the relations used to express the adverbial clauses:

\begin{itemize}
\item \textbf{WHN}: A specific time. 
e.g. "\textit{He listens when you talk.}".
\item \textbf{WHL}: A period of time. 
e.g. "\textit{He listens while you talk.}".
\item \textbf{WHR}: A place. 
e.g. "\textit{He started where he stopped.}".
\item \textbf{IF}: A condition. 
e.g. "\textit{He will come if he can.}".
\item \textbf{SO}: A purpose.
e.g. "\textit{He tries so he can succeed.}".
\item \textbf{BCS}: A reason.
e.g. "\textit{He can't be angry with her because he likes her.}".
\item \textbf{THG}: A concession.
e.g. "\textit{I will come although I don't like traveling.}".
\item \textbf{LIK}: A manner.
e.g. "\textit{He did the job as you asked.}".
\item \textbf{FTR}: After (time).
e.g. "\textit{He start the job after he wake up.}".
\item \textbf{BFR}: Before (time).
e.g. "\textit{Before we came here, the door was shut.}".
\end{itemize}

\subsection{Comparison}

Comparison is a little bit tricky; Mostly it includes an adjective which is shared between two roles, but also to the action. 
For instance, in the sentence "\textit{Karim is taller than his brother}", the adjective "\textit{tall}" is a property of "\textit{Karim}" as it is of "\textit{his bother}". 
We can represent the adjectives in the two roles, but the relation between them (superlative) is better to be represented in the action "To be". 
Sure, the relation of comparison can be represented in one of the roles "\textit{Karim}" and we can reference the other "\textit{his bother}".
But, there are cases where the comparison doesn't include adjectives; it is about the verb instead. 
For example, the sentence "\textit{Karim helps less than his brother}" contains a comparison over the action and not the adjective.
In the end, the comparison must be represented in the action rather than the role.
Table \ref{tab:cmp-eg} contains the two examples quoted previously.
We must point out that the Arabic example for the sentence "Karim helps less than his brother" is not so fluent. 
In order to be fluent, the sentence must be "\RL{kariym 'aqallu musaA`adaTaN min" 'axiyh.} \textit{/Kar\={\i}m aqallu mus\=a`adatan min akh\={\i}h./}".
This can be literally translated as "\textit{Karim is less helpful than his brother}".
In Japanese example, the direct translation is "\textit{Karim is less in help than his brother}". 
Also, we can use "\JP{カリムさんは兄弟よりも少ないながらも手伝いします。} \textit{/Karimu-san wa ky\={o}dai yorimo sukunai nagara tetsudaishimasu./}", which means literally "\textit{Karim helps despite this is less than his brother do.}"; A form which is more polite. 
What matters for us is the comparison itself, when it means "\textit{less than}" we represent it in the "\textbf{cmp}" block.
Then, when it comes to text generation task, each language handles how it has to be generated fluently.

\begin{table}
	\centering
\caption{Example of sentences with comparative.\label{tab:cmp-eg}}%
\begin{tabular}{p{.10\textwidth}p{.85\textwidth}}
\hline\noalign{\smallskip}
Language & Sentence \\
\noalign{\smallskip}\hline\noalign{\smallskip}

Arabic & \RL{kariym 'a.twalu min" 'axiyh.} /Kar\={\i}m a\d{t}walu min akh\={\i}h./ \\
& \RL{kariym yusaA`idu 'aqalla min" 'axiyh.} /Kar\={\i}m yus\={a}`idu aqalla min akh\={\i}h./\\

English & Karim is taller than his brother. \\
& Karim helps less than his brother. \\

French & Karim est plus grand que son frère. \\
& Karim aide moins que son frère. \\

Japanese & \JP{カリムさんは兄弟より背が高いです。} /Karimu-san wa ky\={o}dai yori se ga takai desu./\\
& \JP{カリムさんは兄弟より手伝うのが少ないです。} /Karimu-san wa ky\={o}dai yori tetsudau no ga sukunai desu./\\

\hline\noalign{\smallskip}
\end{tabular}

\end{table}

The idea is to use a block "\textbf{cmp}" in the action block (see Figure \ref{fig:cmp}). 
There are three types of comparison: comparative, superlative and equality. 
If we add \textit{less} comparison, we will have five types: less ("\textbf{L}") and more ("\textbf{M}") for comparative, least ("\textbf{LT}") and most ("\textbf{MT}") for superlative and equal ("\textbf{EQ}").
The first parameter of the comparison would be the agent itself, and the second is a reference in the comparison block.
The STON representations of Table \ref{tab:cmp-eg} examples are illustrated in Figure \ref{fig:cmp-eg}.

\begin{figure}
	\centering
\begin{tabular}{|l|}
\hline
\\
\input{Codes/cmp.tex}
\\
\\
\hline
\end{tabular}
	\caption{Comparison block.}
	\label{fig:cmp}
\end{figure}

\begin{figure}
	\centering
\begin{tabular}{|ll|}
\hline
&\\
\input{Codes/cmp-eg1.tex}
 &  %
\input{Codes/cmp-eg2.tex}
 \\
&\\
Karim is taller than his brother. & Karim helps less than his brother.\\
\hline
\end{tabular}
	\caption{Example of STON representation of comparison.}
	\label{fig:cmp-eg}
\end{figure}

\section{Some issues and their solutions}
\label{sec:issues}

With the Role-Action-Sentence notation, we can represent a lot of sentences. 
But, there are some ambiguities when we want to represent some others. 
To handle this, we will show some problems and their solutions in context of STON.

\subsection{Coordination between references}

One problem is, how to represent the coordination between references. 
There are, principally, two main coordinations which are disjunction ("or") and conjunction ("and"). 
To limit ambiguity when representing a sentence, we want to use either disjunctions of conjunctions or the inverse.
For example:
\begin{itemize}
\item \textit{Mother and son ate food.}
\item \textit{Father and son ate food.}
\end{itemize}
These two sentences can be aggregated in two ways:
\begin{enumerate}
\item \textit{Mother and son \textbf{or} father and son ate food.} (disjunctions of conjunctions);
\item \textit{Mother or father \textbf{and} son ate food.} (conjunctions of disjunctions).
\end{enumerate}
The first one sounds more appropriate than the second.
Moreover, the second sounds like either the mother alone or the father and the son ate the food.
So, the notation of agents and themes must be changed.
Taking the late example, the agents in the sentence can be represented as
"\textbf{agt: [mother, child \textbar\ father, child];}"

\subsection{Proper names}

A proper name is a phrase that identifies one unique entity from a class of entities.
For example, "\textit{London}" is distinguished from the common noun "\textit{city}"; It is more specific.
Proper names can be people (Eg. "\textit{Abdelkrime Aries}"), locations (Eg. "\textit{Algeria}"), organizations (Eg. "\textit{ESI}"), etc.

The problem with proper names is: how to represent them inside the role?
STON is heavily based on Wordnet's synsets (or any other lexicon), each word must have one synset to be represented. 
Some proper names already have synsets in Wordnet, such as U.S. cities. 
But, a city such as "\textit{Jijel}" for example, which is an Algerian city, has no synset number in Wordnet. 
There are many other proper names, such as persons' names, which doesn't exist in Wordnet. 
One solution is to add an attribute "\textbf{nam}" to the role section, which contains the named entity. 

In STON, a role must always have a synset or a pronoun. 
To afford much information about the role (Here, the proper name), we can specify its hypernym's synset.
For instance, the sentence "Karim lives at Jijel" (See Figure \ref{fig:n-e-eg}) has two proper names: "Karim" and "Jijel".
For each, we afford the synset of "person" and "city" respectively.

\begin{figure}
	\centering
\begin{tabular}{|c|}
\hline
\\
\input{Codes/n-e-eg.tex}
\\
\\
\hline
\end{tabular}
	\caption{STON representation of the sentence "\textit{Karim lives at Jijel}".}
	\label{fig:n-e-eg}
\end{figure}

\subsection{Verbs with two objects}

In languages like Arabic and English, a verb can have two objects without using any preposition. 
Table \ref{tab:2obj-eg} represents an example of the sentence "\textit{the man gave the boy a gift}". 
The bold phrase in each language represents the indirect object. 
We can use the prepositions "\textit{to}" and "\RL{li--} \textit{/li/}" with the indirect object.
The English sentence will be "\textit{The man gave a gift \textbf{to the boy}.}". 
Similarly, the Arabic sentence will be "\RL{'a`".taY al-rrajulu hadiyyaTaN \setnashbf lil-.t.tif"li\setnash.} \textit{/a`\d{t}\'{a} al-Rajulu hadiyyatan li-al-\d{T}ifli./}".
So, we can use adpositional phrases to represent the relation with the indirect object. 

\begin{table}
	\centering
\caption{Example of sentences with two objects. \label{tab:2obj-eg}}%
\begin{tabular}{ll}
\hline\noalign{\smallskip}
Language & Sentence \\
\noalign{\smallskip}\hline\noalign{\smallskip}
Arabic & \RL{'a`".taY al-rrajulu \setnashbf al-.t.tif"la \setnash hadiyyaTaN.} \\
& /a`\d{t}\'{a} al-Rajulu al-\d{T}ifla hadiyyatan./  \\
English & The man gave \textbf{the boy} a gift. \\
French & L'homme a donné un cadeau \textbf{à l'enfant}. \\
Japanese & \JP{男性は\textbf{少年に}贈り物があげました。} \\
& /dansei wa sh\={o}nen ni okurimono ga agemashita./ \\
\hline\noalign{\smallskip}
\end{tabular}

\end{table}

\subsection{Pronouns}

In a first time, we thought to use references to the original role instead of personal pronouns; In generation phase, we can generate them when the role is referred many times.
This is can be applied in normal situations when we have all the information including the anaphoric relations. 
In situations when we want to represent sentences generated from extractive summarization for example, we may have a lot of trouble recovering these types of relations.
Also, to boost the analysis (from natural language to STON) and generation (from STON to natural language) tasks, the pronouns are a need.

The pronouns can be classified using many features; Table \ref{tab:pron-class} represents the eight features used for pronouns classification according to \citet{2014-seah-bond}. 
Based on these features, we present the pronouns using two attributes:
\begin{itemize}
\item \textbf{typ:} the type of the pronoun which is encoded on 6 characters:
\begin{itemize}
\item \textbf{1st:} "\textbf{D}" for demonstrative pronoun (this, that, etc.); "\textbf{S}" for subjective personal pronoun (I, he, etc.); "\textbf{O}" for objective personal pronoun (me, him, etc.) and "\textbf{P}" for possessive pronoun (my, his, etc.).
\item \textbf{2nd:} "\textbf{F}" for first person; "\textbf{S}" for second person and "\textbf{T}" for third person.
\item \textbf{3rd:} "\textbf{S}" for singular; "\textbf{D}" for dual; "\textbf{P}" for plural and "\textbf{N}" for not defined number.
\item \textbf{4th:} "\textbf{F}" for female; "\textbf{M}" for male and "\textbf{N}" for neuter. 
Even if the source language attribute sex to objects such as Arabic and French, we consider them as neuter. 
For example, in Arabic, the \textit{chair} is masculine while it is feminine in French.
\item \textbf{5th:} This is reserved for formality and politeness. 
"\textbf{R}" for rude; "\textbf{C}" for casual; "\textbf{F}" for formal and "\textbf{P}" for polite. 
In languages where there is no formality level in pronouns, we choose to use formal by default.
\item \textbf{6th:} The proximity can be: "\textbf{D}" for distal; "\textbf{M}" for medial; "\textbf{P}" for proximal and "\textbf{N}" for not defined.
\end{itemize}
\item \textbf{ref:} the reference(s) to the role(s) related to this pronoun.
\end{itemize}

\setlength{\tabcolsep}{5pt}
\begin{table}
	\centering
\caption{Pronouns features accoding to \citep{2014-seah-bond}.\label{tab:pron-class}}%
\footnotesize
\begin{tabular}{llllllll}
Head & Number & Gender & Case & Type & Formality & Politeness & Proximity  \\
\noalign{\smallskip}\hline\noalign{\smallskip}

Demonstratives  & Dual  & Feminine & Objective & Assertive & Formal & Polite  & Distal \\
Entity  & Plural  & Masculine & Possessive & Elective  & Informal & & Medial \\
Time  & Singular & Neuter  &Subjective & Negative & && Proximal \\
Manner && && Other &&&\\
Person && && Reciprocal &&&\\
Place && && Universal &&&\\
Reason && && Interrogative &&&\\
Thing && && Reflexive &&&\\
Personal (1e, 1i, 2, 3) &&&&&&&\\
Quantifier &&&&&&&\\

\noalign{\smallskip}\hline
\end{tabular}

\end{table}
\setlength{\tabcolsep}{6pt}

The pronouns attribute can be accompanied with a synset to afford a more compact format.
For instance, Figure \ref{fig:pronouns} represents a case where a pronoun and a noun are packed together. 
The pronoun "\textit{his}" is represented as "\textbf{PTSMFN}" which means: possessive, third person, singular, masculine, formal and without proximity.
This representation is better than transforming the clause to "\textit{first novel of him}" before being represented.

\begin{figure}
	\centering
\begin{tabular}{|c|}
\hline
\\
\input{Codes/pronouns.tex}
\\
\\
\hline
\end{tabular}
	\caption{STON representation of the role "\textit{his first novel}".}
	\label{fig:pronouns}
\end{figure}

\subsection{Passive voice}

The passive voice hasn't a specific representation in STON. 
A sentence like "Karim ate the apple" (See Table \ref{tab:passive}) has the same representation as "\textit{The apple was eaten by Karim}". 
The choice of using active or passive voice is decided by the text generation task. 
As for sentences which doesn't have an agent such as "\textit{An apple was eaten}", we simply don't define an agent in the representation.
in this case, when we generate the sentence, we have to use passive voice.

\begin{table}
	\centering
\caption{Example of sentences in passive voice. \label{tab:passive}}%
\begin{tabular}{ll}
\hline\noalign{\smallskip}
Language & Sentence \\
\noalign{\smallskip}\hline\noalign{\smallskip}
Arabic & \RL{al-ttuffaA.haTu 'ukilat" \setnashbf min" .tarafi kariym"\setnash.} \\
& /al-tuffa\d{h}ah ukilat min \d{t}arafi Kar\={\i}m./  \\
English & The apple was eaten \textbf{by Karim}. \\
French & La pomme a été mangée \textbf{par Karim}. \\
Japanese & \JP{林檎が\textbf{カリムさんに}食べられました。} \\
& /Ringo ga Karimu-san ni taberaremashita./ \\
\hline\noalign{\smallskip}
\end{tabular}

\end{table}

\subsection{Complementizer phrase}

A complementizer phrase can be a subject or an object. 
In our representation, we rather consider it as an agent or a theme.
Figure \ref{fig:comp-eg} represents an example of STON annotation in case of complementizer phrases.
Here, the phrase "\textit{STON represents sentences}" is a theme and "\textit{Karim}" is the agent of the action "\textit{Hoping}".

\begin{figure}
	\centering
\begin{tabular}{|l|}
\hline
\\
\input{Codes/comp-eg.tex}
\\
\\
\hline
\end{tabular}
	\caption{Actions of the sentence "Karim hopes that STON represents sentences".}
	\label{fig:comp-eg}
\end{figure}

The structure "\textbf{VERB + TO + INFINITIVE}" will be solved as a complimentizer.
A sentence like "\textit{I want to go there}" can be seen as if "\textit{to go there}" is the theme of "\textit{wanting}".

\section{STON tools and corpora}

STON notation is intended to be used in NLP applications where the concern is about the sentence syntax in a multilingual context.
It can be used in language generation as an intermediate language for machines. 
This implies that we can use it as a mean for text translation, text summarization, etc. 
Parsing STON language would be fast since its grammar is well defined (blocks, references, etc.).
The most challenging task is to create tools to parse a text into STON and to generate a text from STON.
Corpora for testing is a must, this is why we started to annotate some materials already annotated by UNL.

\subsection{Annotation process}

Annotating some materials can be helpful in the future, especially when we intend to use STON for other applications.
Since some grammatical structures are not able to be represented before transformation, like the apposition (which can be replaced by "\textit{which is}").
Here, some steps to follow in order to have a good annotation in STON:
\begin{itemize}
\item Begin to represent roles (Nominal phrases) such as the dependent role must be at last.
For example, in the sentence "the statue of liberty" we represent the role "\textit{liberty}" then the role "\textit{statue}" which has a relation "\textbf{OF}" to the first one.
\item Adjectives can't be represented alone. 
For predicative adjectives, create a role with the noun in the subject and the adjective.
For example, "\textit{The man is friendly}" would be represented as "\textit{The man is friendly man}".
\item As for proper names, if they exist in Wordnet just put their synset (e.g. "\textit{Cairo}"). 
If they don't, put them as value of the attribute "\textbf{nam}" (spaces must be transformed to underscores) and afford the synset of their type: city, person, animal, dog, cat, etc. (the more specific the type, the better).
\item Transform enumerations of a role to a relation "\textbf{OF}". 
For example, the sentence "\textit{He did many jobs: web designer, engineer and teacher}" will be "\textit{He did many jobs \textbf{of} web designer, engineer and teacher}".
\item When you find some prepositions we didn't talk about, try to find a similar in the relations. 
For example, The preposition "\textit{As well as}" can be considered as a simple "\textit{and}".
\item Omit the chronological order indicators (first, then, finally) and represent it as consecutive actions.
\item Try to transform expressions in order to fit the STON representation.
For example: "\textit{much of it experimental}" will be "\textit{which is much experimental}".
\item Comparison doesn't exist in roles, it just exists in actions. 
To solve this problem, we add the expression "\textit{an amount which is }"; the comparison must not have an adjective.
For example, "\textit{the author of more than 200 articles}" will  be "\textit{the author of an amount which is more than 200 articles}".
\end{itemize}

We started to annotate some biographies\footnote{\url{http://www.undl.org/unldoc/bb.htm}} already annotated by UNL.
You can check the annotated texts on Nolporas project\footnote{Nolporas project: \url{https://github.com/kariminf/NaLanPar}} which aims to create corpora mostly for STON.
Table \ref{tab:corpora} represents some statistics on the two biographies we annotated.
The biographies contains long sentences with an average of 29 words.
The synsets in STON are the equivalent of UWs in UNL; here the number is less because we use references to roles.
Roles sometimes are repeated in other sentences, so we don't need to repeat their descriptions.
Relations in UNL corresponds to the relations (adpositions, relatives and adverbials) and the references of themes and agents.
Concerning the manual annotation, each sentence took us from a half to one hour since we have to choose between different senses and the right relations.

\begin{table}
	\centering
\caption{Statistics on the annotated texts. \label{tab:corpora}}%
\begin{tabular}{p{.25\textwidth}p{.25\textwidth}p{.25\textwidth}}
\hline\noalign{\smallskip}
 & Naguib Mahfouz Bio &  Louis de Broglie Bio \\
\noalign{\smallskip}\hline\noalign{\smallskip}

\multicolumn{3}{c}{Original texts statistics: } \\
\noalign{\smallskip}

\# Sentences & 10 & 35 \\
\# Words & 287 & 1055 \\
\# Words/Sentence & 29 & 30 \\

\noalign{\bigskip}

\multicolumn{3}{c}{UNL statistics: } \\
\noalign{\smallskip}
\# UWs with redundancy & 264 & 803 \\
\# Relations & 172 & 507 \\
\# Attributes & 229 & 703\\

\noalign{\bigskip}

\multicolumn{3}{c}{STON statistics: } \\
\noalign{\smallskip}

\# Synsets with redundancy & 132 & 373 \\
\# Roles & 85 & 237 \\
\# Actions & 23 & 82 \\
\# Relations & 61 & 183 \\

\noalign{\bigskip}
\hline
\end{tabular}

\end{table}

The annotation process must be automatic which will be more interesting. 
We started working on another project called NaLanPar\footnote{NaLenPar source: \url{https://github.com/kariminf/NaLanPar}} (Natural language parser) which aims to generate sentence representations, including STON, from texts.
It uses open source text parsers such as Stanford Parser \citep{2014-chen-manning}, which is a syntactic parser for many languages (English, Arabic, etc.) licensed under GPL license. 
Currently, we are working on transforming English text to STON, and more languages are to be added after finishing. 
At the mean time, it can handle the sentences of form "Subject Verb Object Preposition Noun".

\subsection{STON parsing}

We created a parser for STON in a project called SentRep\footnote{SentRep source: \url{https://github.com/kariminf/SentRep}}(sentence representation).
Its aim is to implement parsers for different sentence representation languages, taking STON as primary focus.
The parser is an abstract class which can be extended to implement the cases where the parser finds a role, adjective, action, etc. 

To test the speed of parsing, we used a PC with i7 processor (2 GHz 4 cores) and openJDK 7. 
The parser do nothing when it finds the different parts (just calls the functions which will treat them).
For each biography (NaguibMahfouz and LouisdeBroglie) we executed the parser 10 times and calculated the time of parsing in milliseconds.
NaguibMahfouz biography parsing took between 60ms and 71ms with an average of 67ms, and almost 8ms per sentence.
As for LouisdeBroglie biography parsing, it took between 132ms to 176ms with an average of 163ms, and almost 5ms per sentence.

\subsection{Generation}

To generate sentences from STON representation, we are currently working on another project called NaLenGen\footnote{NaLenGen source: \url{https://github.com/kariminf/NaLanGen}} (Natural language generator).
NaLenGen aims to generate text from different sentence representations (It uses SentRep for that). 
It uses open source text realizers such as SimpleNLG \citep{2009-gatt-reiter}, which is a realization engine for English licensed under MPL license. 
It has been adapted to many other languages: German \citep{2011-bollmann}, French \citep{2013-vaudry-lapalme} and Brazilian Portuguese \citep{2014-deoliveira-al}.
To map Wordnet synsets to other languages, we use Open Multilingual Wordnet \citep{2013-bond-foster}. 
The mapping is not complete for many languages, as a result the generation may fail when the synset is not found.

We tried to generate some English and French text from STON annotation. 
Examine Table \ref{tab:generate} which is an example of the two first sentences of Naguib Mahfouz biography.
The resulted text was fair for English and a little bad for French.
This is why more work has to be done to address some issues we have found:
\begin{itemize}
\item Sometimes the mapping to other languages is done automatically which leads to some errors in translating concepts. 
For example, we found that the concept "\textit{Cairo}" was mapped to "\textit{OSS 117 : Le Caire}" in French.
As we can see in the example, many other concepts are wrong.
\item The synsets contain many words, and some words are more adequate than others.
This is why we have to propose a method in order to choose these words based on their frequency of use for example.
\item We can't generate directly from STON, and some changes must be done to have more fluent texts. 
For example, predicative adjectives are expressed as a role with a noun.
The sentence "\textit{The child is happy}" is represented as "\textit{The child is happy child}".
\end{itemize}

\begin{table}
	\centering
\caption{Example of sentences generation to English and French.\label{tab:generate}}%
\begin{tabular}{p{.13\textwidth}p{.8\textwidth}}
\hline\noalign{\smallskip}
Source & 
Born in Cairo in 1911, Naguib Mahfouz began writing when he was seventeen. 
His first novel was published in 1939 and ten more were written before the Egyptian Revolution of July 1952, when he stopped writing for several years.
\\
\noalign{\smallskip}

English generated text & 
Naguib Mahfouz which was given birth in Cairo in 1911 began writing when he was 17 years. 
First, his novel was published in 1939 and 10, more novels were written before the revolution of Egyptians of July 1952 in which he discontinued writing for several years. 
\\
\noalign{\smallskip}

French generated text & 
Naguib Mahfouz que a été accouché à un Le Caire à 1911 a débuté un œuvre quand lui a été de 17 années. 
Son premier nouveau a été publié à 1939 et de 10 nouveaux plus ont été écrits avant le tour de des égyptiens de July 1952 à lequel lui a cessé un œuvre pour des années es. 
\\
\noalign{\smallskip}\hline

\end{tabular}

\end{table}

\section{Limitations and challenges}

Although STON can represent a wide range of sentences, it shows some limitations. 
Sometimes, the variation between languages can prevent us from representing the sentences properly. 
In some languages, we don't always find the same syntactic representation of a meaning. 
For instance, considering Table \ref{tab:hungry}, the adjective "\textit{hungry}" has no exact translation in Japanese. 
In fact, the Japanese sentence literally means "my stomach emptied". 
The same adjective is translated to a noun in French, where the sentence literally means "\textit{I have a hunger}".
In Arabic, we can use the sentence "\RL{'anA ju`t.} \textit{/an\=a ju`t/}", which means the same thing. 
This sentence is composed of the pronoun "\textit{I}" and a verb "\textit{to be hungry}" conjugated in the past. 
Also, since STON is based on Wordnet synsets, it is limited to the concepts extracted from English language. 
So, if we want to represent a sentence from Arabic for example, we have to find Wornet concepts that are close to the meaning of the sentence's words.

STON can't deal with text structures like paragraphs, lists, tables, etc. 
Many languages have complex predicates which can represent the same meaning as a singleton verb. 
For instance, "\textit{I gave the baby a bath}" and "\textit{I bathed the baby}" have the same meaning. 
Unfortunately, STON is not a fully semantic representation which means it handles those as two different forms.
A sentence like "\textit{the purpose is to teach mathematics and develop physics}" can be represented without problem.
But, in case of this sentence: "\textit{the purpose is to teach and develop physics}", we have to use redundancy and represent it as "\textit{the purpose is to teach physics and develop physics}".

\begin{table}
	\centering
\caption{Example of sentences which doesn't have the same POS in different languages. \label{tab:hungry}}%
\begin{tabular}{lll}
\hline\noalign{\smallskip}
Language & Sentence & Romanization\\
\noalign{\smallskip}\hline\noalign{\smallskip}
Arabic & \RL{'anA jaw`aAn.} 
& /an\={a} jaw`\={a}n./ \\
English & I am hungry. &\\
French & J'ai faim. &\\
Japanese & \JP{お腹が空いた。} 
& /onaka ga suita./ \\
\noalign{\smallskip}\hline
\end{tabular}

\end{table}

\section{Discussion}
\label{sec:discussion}

STON is all about representing the different parts of a sentence independently from its structure in natural languages. 
It is meant to transport sentences information between different applications (programs). 
It represents the syntactic relations between the different parts of sentences in a multi-lingual way.
When the syntax fails to keep the multilingual aspect, it uses semantic relations instead.
To understand STON better, we have to know what this language is not about. 
\begin{itemize}
\item It does not represent the relations between the parts of speech semantically, even if there are some relations in relative clauses. 
For example, it does not allow us to represent relations like UNL does, such as beneficiary relation and purpose relation.
\item It is not a format for storing texts, such as Open document format (ODF) and Microsoft office format which are based on XML. 
\end{itemize}
There are similarities and differences between STON and other representation languages.
A concise comparison between KANT, UNL, AMR and STON is given in Table \ref{tab:ann-comp}.

\begin{table}
	\centering
\caption{Comparaison between STON and other well-known annotation formats. \label{tab:ann-comp}}%
\begin{tabular}{p{.13\textwidth}p{.18\textwidth}p{.18\textwidth}p{.18\textwidth}p{.18\textwidth}}
\hline\noalign{\smallskip}
Criteria & KANT & UNL & AMR & STON  \\
\noalign{\smallskip}\hline\noalign{\smallskip}
Objective 
& Interlingua for machine translation from English technical manuals.
& Represent the meaning of texts without ambiguity, to be used as a language of the web.
& Write down the meanings of English sentences.
& Represent sentences in a multilingual way without basing so much on semantics, 
to be used as interchange format between applications.
\\
\noalign{\smallskip}

Aspects 
& Semantics, with morphological aspects (tense, etc.) 
& Morphology, Semantics,  Pragmatics
& Semantics, no morphological aspects (tense, etc.)
& Syntax, with morphological aspects (tense, etc.) and semantics when syntax fails to be multilingual
\\
\noalign{\smallskip}

Concepts
& Own
& UWs (UNL Ontology)
& PropBank frames
& WordNet synsets
\\
\noalign{\smallskip}

Dependency
& Domain dependent
& Language independent
& English language dependent
& Language independent
\\
\noalign{\smallskip}

Relations
& Own
& Own
& PropBank relations
& Own
\\
\noalign{\smallskip}

Readability 
& Readable 
& Difficult to follow
& Less readable
& Less readable, difficult when we have a big text
\\
\noalign{\smallskip}
\hline
\end{tabular}

\end{table}

When we check the meaning, UNL is the representation that mostly represents the meaning. 
AMR, in the other hand, is used to represent the meaning but it lacks some morphological aspects such as verb tense.
KANT and STON annotations are less depending on semantics than the two previous ones. 
Our objective is to represent sentences with a minimum cost (time and processing effort).
This is why we represent relations like "\textbf{from}" as they are, even if they can mean many things: "\textit{from 2 am}", "\textit{from London}", etc.

Semantic representation is a powerful tool because it allows us to represent what could be understood from a sentence. 
It can eliminate redundancy in sentences; For example \citep{2013-banarescu-al}:
\begin{itemize}
\item \textit{The soldier was afraid of battle.}
\item \textit{The soldier feared battle.}
\item \textit{The soldier had a fear of battle.}
\end{itemize}
These sentences have the same meaning, therefore their representation must be the same.
In contrast with UNL and AMR, STON doesn't go deep into the semantic relations, or to comprehend the sentence as a whole.
Figure \ref{fig:ston-sem} shows STON representation of these 3 sentences. 
It is clear that the representation makes difference between the verbs, nouns and adjectives.

\begin{figure}
	\centering
\begin{tabular}{|c|}
\hline
\\
\input{Codes/ston-sem.tex}
\\
\\
\hline
\end{tabular}
	\caption{STON action representation of 3 sentences with same meaning.}
	\label{fig:ston-sem}
\end{figure}

The four languages are based on different ontologies and lexicons to represent their concepts.
KANT annotation is based on concepts defined especially for the KANT system, generally limited by technical reports domain.
Likewise, UNL defines its own concepts base called UNL ontology, and each concept is referred to as a universal word (UW). 
AMR and STON use PropBank and Wordnet respectively, this is why they are limited to these two bases.

AMR and KANT are more based on English, while UNL and STON seeks to be multilingual. 
When it comes to multilingual aspect, the UWs of UNL are so powerful. 
For the mean time, we use Wordnet to represent the different concepts. 
Unfortunately, till nowadays, the mapping to other languages is not complete.
For this reason, we had some problems in generating French text from STON annotation when the synset can't be found.

Readability is important when we want to create a sentence representation manually or to check it after automatic generation.
KANT annotation is more readable than the other three representations.
STON is developed as a machine language such as UNL, even so, we want to allow some space for readability.
This can be helpful in case we want to test a system that uses STON, because it will be easier to create test banks.

\section{Conclusion}
\label{sec:conclusion}

In this work, we proposed an inter-application language (STON) aimed to represent sentences structures in a multilingual context. 
STON is based somehow on the JSON representation with some adjustments to speed up the parsing.
It is based on the assumption that anything in the sentence is either a role or an action with relations between them. 
The representation uses both syntactic structure (noun definition, verb tense, subjects, objects, etc) and semantic relations (time and place relations, etc.). 
To support multilingualism, we use concepts instead of words (in our case, we use Wordnet's synsets).
Our intention is to use STON as a mean of communication between different applications.
More specifically, the language is intended to be used in cross-lingual automatic text summarization.

STON is far beyond being complete or being perfect.
There still are some improvements to be made in the future, such as the problem of words' syntactic alignment inter-languages.
Because it is based on Wordnet, the concepts are limited to English. 
There are a lot of concepts that doesn't exist in English but exist in other languages. 
Exploiting a larger semantic network with a more knowledge base like BabelNet\footnote{\url{http://babelnet.org}} \citep{2012-navigli-ponzett} may improve inter-languages representation.

\section*{Acknowledgement}
Special thanks to Hisham Omar for his valuable feedback concerning Japanese examples.

\bibliographystyle{unsrtnat}
\bibliography{ston}

\begin{thebibliography}{21}
\providecommand{\natexlab}[1]{#1}
\providecommand{\url}[1]{\texttt{#1}}
\expandafter\ifx\csname urlstyle\endcsname\relax
  \providecommand{\doi}[1]{doi: #1}\else
  \providecommand{\doi}{doi: \begingroup \urlstyle{rm}\Url}\fi

\bibitem[Bertran et~al.(2008)Bertran, Borrega, Recasens, and
  Soriano]{2008-bertran-al}
Manuel Bertran, Oriol Borrega, Marta Recasens, and B\`{a}rbara Soriano.
\newblock Ancorapipe: A tool for multilevel annotation.
\newblock 41:\penalty0 291--292, 2008.
\newblock ISSN 1135-5948.

\bibitem[Recasens and Mart\'{\i}(2009)]{2009-recasens-marti}
Marta Recasens and M.~Ant\`{o}nia Mart\'{\i}.
\newblock Ancora-co: Coreferentially annotated corpora for spanish and catalan.
\newblock \emph{Language Resources and Evaluation}, 44\penalty0 (4):\penalty0
  315--345, 2009.
\newblock ISSN 1574-0218.
\newblock \doi{10.1007/s10579-009-9108-x}.
\newblock URL \url{http://dx.doi.org/10.1007/s10579-009-9108-x}.

\bibitem[Lopatkov{\'a} et~al.(2011)Lopatkov{\'a}, Homola, and
  Klyueva]{2011-lopatkova-al}
Mark{\'e}ta Lopatkov{\'a}, Petr Homola, and Natalia Klyueva.
\newblock Annotation of sentence structure.
\newblock \emph{Language Resources and Evaluation}, 46\penalty0 (1):\penalty0
  25--36, 2011.
\newblock \doi{10.1007/s10579-011-9162-z}.
\newblock URL \url{http://dx.doi.org/10.1007/s10579-011-9162-z}.

\bibitem[Miller(1995)]{1995-miller:1995}
George~A. Miller.
\newblock Wordnet: A lexical database for english.
\newblock \emph{Commun. ACM}, 38\penalty0 (11):\penalty0 39--41, November 1995.
\newblock ISSN 0001-0782.
\newblock \doi{10.1145/219717.219748}.
\newblock URL \url{http://doi.acm.org/10.1145/219717.219748}.

\bibitem[Uchida et~al.(1999)Uchida, Zhu, and Della~Senta]{1999-uchida-al}
Hiroshi Uchida, Meiying Zhu, and Tarcisio Della~Senta.
\newblock The unl, a gift for a millennium.
\newblock 1999.
\newblock URL \url{http://www.undl.org/publications/gm/index.htm}.

\bibitem[Banarescu et~al.(2013)Banarescu, Bonial, Cai, Georgescu, Griffitt,
  Hermjakob, Knight, Koehn, Palmer, and Schneider]{2013-banarescu-al}
Laura Banarescu, Claire Bonial, Shu Cai, Madalina Georgescu, Kira Griffitt, Ulf
  Hermjakob, Kevin Knight, Philipp Koehn, Martha Palmer, and Nathan Schneider.
\newblock Abstract meaning representation for sembanking.
\newblock In \emph{Proceedings of the 7th Linguistic Annotation Workshop and
  Interoperability with Discourse}, pages 178--186, Sofia, Bulgaria, August
  2013. Association for Computational Linguistics.
\newblock URL \url{http://www.aclweb.org/anthology/W13-2322}.

\bibitem[Mitamura et~al.(1991)Mitamura, Nyberg, and
  Carbonell]{1991-mitamura-al}
Teruko Mitamura, Eric~H Nyberg, and Jaime~G Carbonell.
\newblock An efficient interlingua translation system for multi-lingual
  document production.
\newblock In \emph{Proceedings of the Third Machine Translation Summit}, 1991.

\bibitem[Czuba et~al.(1998)Czuba, Mitamura, and Nyberg]{1998-mitamura-al}
Krzysztof Czuba, Teruko Mitamura, and Eric~H Nyberg.
\newblock Can practical interlinguas be used for difficult analysis problems?
\newblock In \emph{Proceedings of AMTA-98 Workshop on Interlinguas}, 1998.

\bibitem[Uchida and Zhu(2005)]{2005-uchida-zhu}
Hiroshi Uchida and Meiying Zhu.
\newblock Unl2005 from language infrastructure toward knowledge infrastructure.
\newblock \emph{Special Speech, Pacific Association for Computational
  Linguistics (PCLING 2005)}, 2005.

\bibitem[Boguslavsky(2013)]{2005-boguslavsky}
Igor Boguslavsky.
\newblock Some lexical issues of unl.
\newblock \emph{Universal Networking Language: Advances in Theory and
  Applications}, pages 101--108, 2013.

\bibitem[Martins(2013)]{2013-martins}
R.~Martins.
\newblock \emph{Lexical Issues of UNL: Universal Networking Language 2012
  Panel}.
\newblock EBSCO ebook academic collection. Cambridge Scholars Publishing, 2013.
\newblock ISBN 9781443852814.
\newblock URL \url{https://books.google.dz/books?id=tdcwBwAAQBAJ}.

\bibitem[Kingsbury and Palmer(2002)]{2002-kingsbury-palmer}
Paul Kingsbury and Martha Palmer.
\newblock From treebank to propbank.
\newblock In \emph{Language Resources and Evaluation}, 2002.

\bibitem[Haak(1997)]{1997-haak}
M.~Haak.
\newblock \emph{The Verb in Literary and Colloquial Arabic}.
\newblock Functional grammar series. Mouton de Gruyter, 1997.
\newblock ISBN 9783110154016.
\newblock URL \url{https://books.google.dz/books?id=I21G6qVQibkC}.

\bibitem[Seah and Bond(2014)]{2014-seah-bond}
Yu~Jie Seah and Francis Bond.
\newblock Annotation of pronouns in a multilingual corpus of mandarin chinese,
  english and japanese.
\newblock In \emph{10th Joint ACL - ISO Workshop on Interoperable Semantic
  Annotation}, pages 82--87, Reykjavik, Iceland, 2014.

\bibitem[Chen and Manning(2014)]{2014-chen-manning}
Danqi Chen and Christopher Manning.
\newblock A fast and accurate dependency parser using neural networks.
\newblock In \emph{Proceedings of the 2014 Conference on Empirical Methods in
  Natural Language Processing (EMNLP)}, pages 740--750, Doha, Qatar, October
  2014. Association for Computational Linguistics.
\newblock URL \url{http://www.aclweb.org/anthology/D14-1082}.

\bibitem[Gatt and Reiter(2009)]{2009-gatt-reiter}
Albert Gatt and Ehud Reiter.
\newblock {SimpleNLG}: A realisation engine for practical applications.
\newblock In \emph{Proceedings of the 12th European Workshop on Natural
  Language Generation (ENLG 2009)}, pages 90--93, Athens, Greece, March 2009.
  Association for Computational Linguistics.
\newblock URL \url{http://www.aclweb.org/anthology/W09-0613}.

\bibitem[Bollmann(2011)]{2011-bollmann}
Marcel Bollmann.
\newblock Adapting simplenlg to german.
\newblock In \emph{Proceedings of the 13th European Workshop on Natural
  Language Generation}, pages 133--138, Nancy, France, September 2011.
  Association for Computational Linguistics.
\newblock URL \url{http://www.aclweb.org/anthology/W11-2817}.

\bibitem[Vaudry and Lapalme(2013)]{2013-vaudry-lapalme}
Pierre-Luc Vaudry and Guy Lapalme.
\newblock Adapting simplenlg for bilingual english-french realisation.
\newblock In \emph{Proceedings of the 14th European Workshop on Natural
  Language Generation}, pages 183--187, Sofia, Bulgaria, August 2013.
  Association for Computational Linguistics.
\newblock URL \url{http://www.aclweb.org/anthology/W13-2125}.

\bibitem[de~Oliveira and Sripada(2014)]{2014-deoliveira-al}
Rodrigo de~Oliveira and Somayajulu Sripada.
\newblock Adapting simplenlg for brazilian portuguese realisation.
\newblock In \emph{Proceedings of the 8th International Natural Language
  Generation Conference (INLG)}, pages 93--94, Philadelphia, Pennsylvania,
  U.S.A., June 2014. Association for Computational Linguistics.
\newblock URL \url{http://www.aclweb.org/anthology/W14-4412}.

\bibitem[Bond and Foster(2013)]{2013-bond-foster}
Francis Bond and Ryan Foster.
\newblock Linking and extending an open multilingual wordnet.
\newblock In \emph{Proceedings of the 51st Annual Meeting of the Association
  for Computational Linguistics (Volume 1: Long Papers)}, pages 1352--1362,
  Sofia, Bulgaria, August 2013. Association for Computational Linguistics.
\newblock URL \url{http://www.aclweb.org/anthology/P13-1133}.

\bibitem[Navigli and Ponzetto(2012)]{2012-navigli-ponzett}
Roberto Navigli and Simone~Paolo Ponzetto.
\newblock {B}abel{N}et: {T}he automatic construction, evaluation and
  application of a wide-coverage multilingual semantic network.
\newblock \emph{Artificial Intelligence}, 193:\penalty0 217--250, 2012.

\end{thebibliography}

\end{document}